\documentclass[10pt,twocolumn,letterpaper]{article}

\usepackage{cvpr}              

\usepackage{graphicx}
\usepackage{amsmath}
\usepackage{amssymb}
\usepackage{booktabs}
\usepackage{graphicx}
\usepackage{amsmath}
\usepackage{amssymb}
\usepackage{booktabs}
\usepackage{comment}
\usepackage{graphicx}
\usepackage{subcaption}
\usepackage{float}

\usepackage{textcomp}
\usepackage{cvpr}
\usepackage{times}
\usepackage{epsfig}
\usepackage{mathtools, nccmath}
\usepackage{float}
\usepackage{longtable}
\usepackage{multirow}
\usepackage[utf8]{inputenc}
\usepackage[spanish,english]{babel}
\usepackage[table,xcdraw]{xcolor} 

\usepackage[breaklinks=true,bookmarks=false]{hyperref}

\usepackage[capitalize]{cleveref}
\crefname{section}{Sec.}{Secs.}
\Crefname{section}{Section}{Sections}
\Crefname{table}{Table}{Tables}
\crefname{table}{Tab.}{Tabs.}

\def\cvprPaperID{*****} 
\def\confName{CVPR}
\def\confYear{2022}

\begin{document}

\title{Classification of Chest XRay Diseases through image processing and analysis techniques}

\author{
  Santiago Martínez Novoa\textsuperscript{1}\\
  {\tt\small s.martinezn@uniandes.edu.co}
  \and
  María Catalina Ibáñez\textsuperscript{1}\\
  {\tt\small m.ibanez@uniandes.edu.co}
  \and
  Lina Gómez Mesa\textsuperscript{1}\\
  {\tt\small l.gomez1@uniandes.edu.co}
  \and
  Jeremias Kramer\textsuperscript{1}\\
  {\tt\small j.kramer@uniandes.edu.co}
  \\ \textsuperscript{1}Universidad de los Andes
}

\maketitle

\begin{abstract}
    Multi-Classification Chest X-Ray Images are one of the most prevalent forms of radiological examination used for diagnosing thoracic diseases. In this study, we offer a concise overview of several methods employed for tackling this task, including DenseNet121. In addition, we deploy an open-source web-based application. In our study, we  conduct tests to compare different methods and see how well they work. We also look closely at the weaknesses of the methods we propose and suggest ideas for making them better in the future. Our code is available at: \url{https://github.com/AML4206-MINE20242/Proyecto_AML}
\end{abstract}

\section{Introduction}
\label{sec:intro}

Interpreting chest radiographs (CXR) remains a significant challenge in contemporary clinical practice, particularly concerning the diagnosis of lung diseases such as pneumonia, tuberculosis, and chronic obstructive pulmonary disease (COPD). Despite advancements in medical diagnostics, the complexity of the thorax and the subtlety of disease markers make CXR analysis a nuanced and time-consuming process. Moreover, the scarcity of radiological expertise, especially in underserved regions, exacerbates the challenge, potentially delaying diagnoses and treatment interventions for millions of patients worldwide.\cite{Majkow}

The inherent ambiguity of CXR presentations poses a diagnostic hurdle even for experienced healthcare professionals. Factors such as patient positioning, reader expertise, and environmental conditions can influence interpretation, leading to diagnostic errors. Additionally, the shortage of trained radiologists further compounds the burden of CXR interpretation, contributing to delayed diagnoses and impacting patient outcomes.

In response to these challenges, Computer-Aided Diagnosis (CAD) systems leveraging Artificial Intelligence (AI) and Computer Vision (CV) present a promising solution. By harnessing techniques like convolutional neural networks (CNNs) and transfer learning, these systems aim to enhance the accuracy and efficiency of disease classification from CXRs. The primary objective is to alleviate the burden on radiologists, improve diagnostic accuracy, and expedite treatment interventions. \cite{MEEDENIYA2022109319}

In this paper, we aim to explore methods for automated chest X-ray interpretation, focusing on the training of algorithms to identify various lung diseases. By addressing the aforementioned challenges and leveraging advancements in deep learning and computer vision, we endeavor to improve healthcare outcomes by providing advice and detailed suggestions to help healthcare professionals in their diagnoses. Through multi-class and multi-label classification, image enhancement and segmentation our objectives include providing accurate and timely diagnoses while ensuring model explainability and usability in real-world clinical settings.

\textbf{Objectives:} Our objectives are as follows:

\begin{enumerate}

    \item Develop an advanced algorithm capable of accurately detecting and classifying various types of lung diseases from chest X-ray images.
    
    \item Enhance operational efficiency by automating aspects of the disease identification process, serving as a complementary tool to radiological expertise rather than a replacement.
    
    \item Improve patient outcomes by providing timely support to healthcare professionals through the algorithm's ability to identify deterministic characteristics and offer a valuable second opinion during the diagnostic process.

\end{enumerate}
\section{State of the art}
\label{sec:formatting}

\subsection{Label powerset and classification chains}

Allaouzi et al. \cite{Stateoftheart_LR} proposed a novel approach for multi-label chest x-ray classifications of common thorax diseases. The experimental study was tested on two publicly available CXR datasets: ChestX-ray14 (frontal view) and CheXpert (frontal and lateral views).Their proposed approach consists of four parts: data description and exploration, data pre-processing, feature extraction and classification. 

For the CheXpert dataset they worked only with 134,327 CXRs (115,723 of frontal views and 18604 of lateral views), because they ignored CXRs with uncertain  labels. 

They resized CXRs to the required size 224x224 pixels. They also used horizontal flipping and normalized their data by substracting the mean from each pixel and divided the results by the standard deviation to augment the dataset and make convergence faster. 

The main goal of this phase was to generate the features that were fed to the classifier to classify the CXR into one or more possible classes. To this end, Allaouzi et al. \cite{Stateoftheart_LR} used a denseNet-121 model. 

To classify the CXRs they used problem transformation methods such as Binary Relevance (BR), Label PowerSet (LP) and Classifier Chains (CC). The base classifier algorithm used with these methods was Logistic Regression (LR).

They used CC and LP to exploit the correlation between pathologies (likelihood that the presence of certain pathology could determine if another is also present or not) and this way overcome the label independence assumption of BR. However, these methods did not give them desired results because they weren't significantly better compared to BR. This indicates the use of a specific transformation method not necessarily improves the performance of the model and so it would be beneficial to try adaptation or ensemble methods.

\subsection{Self-supervised training and Binary Mapping}

Irvin et al. \cite{irvin2019chexpert} investigated different approaches to predict the probability of 14 different observations from multi-view chest radiographs using the CheXpert dataset. Three main approaches were explored to handle uncertainty in training labels: Ignoring, Binary Mapping, Self-Training and 3-Class Classification.

This baseline approach ignores uncertain labels during training. It optimizes binary cross-entropy loss, akin to listwise deletion in imputation methods. However, this approach ignores a significant portion of data, that could potentially lead to biased models.

In this approach uncertain labels are replaced with either 0 or 1. It mimics zero imputation strategies but may distort classifier decision-making if uncertainty carries meaningful information.

Finally, in this approach uncertain labels are treated as unlabeled examples, employing semi-supervised learning. After training a model without uncertainty labels, predictions are used to relabel uncertain instances. However a disadvantage of this approach can be the reliance on model predictions because it may introduce error propagation. 

Additionally, the study explores a 3-class classification approach, treating uncertainty as its own class. This method allows better incorporation of uncertainty information, potentially enhancing network understanding of different pathologies. 

All approaches are trained using DenseNet121 architecture.

\subsection{Conditional training and label smoothing regularization}

Pham et al. \cite{pham2020interpreting} present a supervised multi-label classification framework based on deep convolutional neural networks (CNNs) for predicting the presence of 14 common thoracic diseases and observations using CheXpert dataset CXR images. Uncertainty in training labels is addressed through a novel training procedure and leveraging label smoothing regularization (LSR).

The network architecture used in the study is DenseNet-121, chosen as the baseline model. The final layer of the network consists of a 14-dimensional dense layer followed by sigmoid activations, enabling the prediction of probabilities for each of the 14 pathology classes present in the chest X-ray (CXR) images.

Ablation studies were performed to evaluate the effectiveness of conditional training (CT) and label smoothing regularization (LSR). Initially, the baseline network is trained independently using different label approaches: U-Ignore, U-Ones, and U-Zeros. Subsequently, CT and LSR are applied individually and in combination to assess their impact on model performance.

Lastly, to address variability in AUC scores across different architectures, a model ensemble strategy is employed. Six state-of-the-art CNN models, including DenseNet-169, DenseNet-201, Inception-ResNet-v2, Xception, and NASNetLarge, are trained and evaluated alongside DenseNet-121. The ensemble model averages the outputs of all trained networks, and test-time augmentation (TTA) is applied to enhance prediction accuracy.

\section{Approach and Metrics}
\subsubsection{Approach}

\begin{figure}[H]
  \centering

  {\includegraphics[width=1\linewidth]{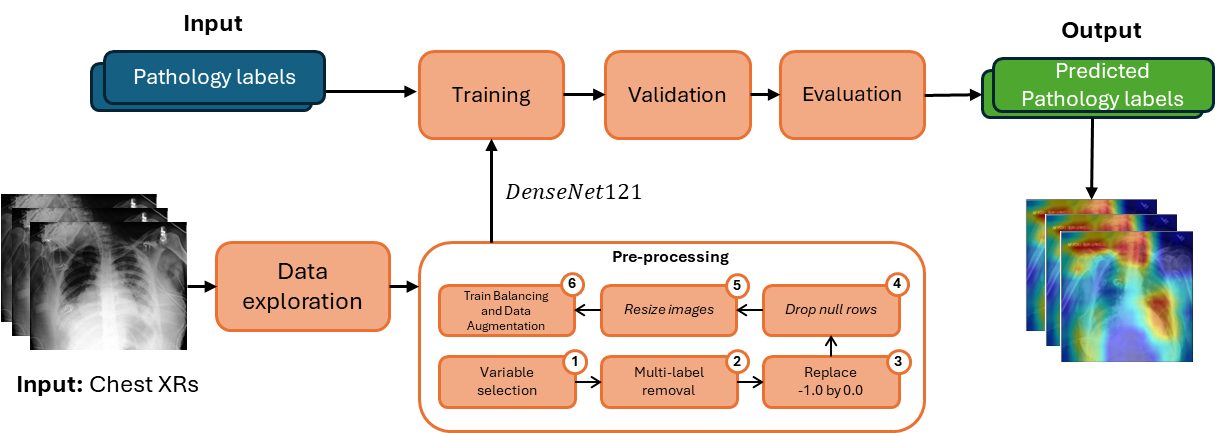}}%
  \hfill

  \caption{Final approach for multi-classification task on the CheXPert Dataset}
  \label{fig:img1}
\end{figure}

The proposed method consists of 4 steps: 1) data description and exploration, 2) data pre-processing, 3) training of a 5-class DenseNet121, EfficientNet121 and ViT pretrained models to classify chest X-ray images with No Finding, Cardiomegaly, Edema, Pneumotorax and Pleural Effusion and 4) an evaluation of the performance of the models, as it can be seen in \ref{fig:img1}. In the data exploration step, an in-depth analysis of the dataset was conducted to gain insights into each class characteristics and distribution. This step is further described in the dataset section of this paper. 

In second place, in the pre-processing step, the focus was to prepare the dataset to train the model. This step involved filtering the data to only contain the chosen classes, checking the labels were in the correct format, dropping rows where all attributes were completely null, resizing the images and applying transformations, like horizontal and vertical flipping as it is the most common data augmentation technique to balance datasets in the medical domain \cite{subra}. Data augmentation increases the scope of the training data by creating different repetitions of the data, often applied to prevent overfitting by acting as a regularizer and facilitating model convergence. The data enhancement allows the model to generalize features through exposure to different variations of the image \cite{ukwan}.

Finally, after applying all the pre-processing steps, we varied the model used to classify. We decided to vary the model to ensure the best possible classification results. By experimenting with multiple models, we could identify the one that best captured the underlying patterns in our data to avoid the most overfitting. We also changed the learning rate, as this hyperparameter significantly influences how quickly and effectively a model learns from the data. Adjusting the learning rate allowed us to fine-tune the training process and potentially achieve even better performance by balancing speed and accuracy. Some common values used in literature for the learning rate in medical imaging classification are 1e-2 and 1e-4 \cite{shlapentokh2024exploring}.

\subsubsection{Architectures}

We propose a Convolutional Neural Network architecture inspired by DenseNet121  \cite{HuangLW16a} as the backbone and as our baseline model, where it receives the input image $x_0$ of size (1,224,224) and passes it through a sequence of dense and transitional layers. During this process, the input goes through a dimensionality reduction process (downsampling) until it reaches the final  layer. At this stage, the image has been transformed into a lower dimensional space. Then, a global average pool layer and a fully connected layer with softmax activation function with the respective predicted label are presented.

\begin{figure*}[]
  \centering

  {\includegraphics[width=0.95\linewidth]{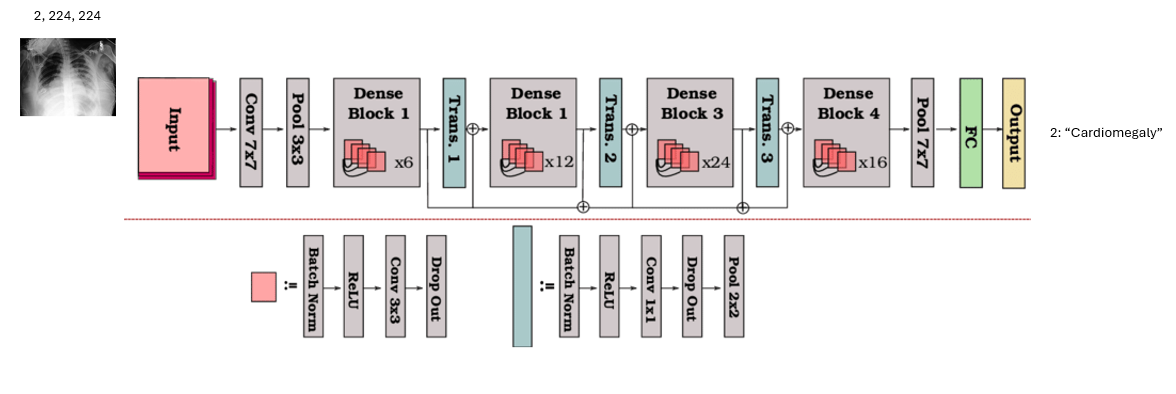}}%
  \hfill

  \caption{Baseline architecture that consists of 4 dense blocks}
  \label{fig:baseline}
\end{figure*}

The overview architecture of our proposed framework is shown in Fig. \ref{fig:baseline}. Given a grayscale input image $x_{L} \in \mathbb{R} ^ {H \times W \times 1}$, the image first passes through the dense blocks, which consist of convolutional blocks with batch normalization, ReLU, 3x3 Convolutions, and dropout layers. Each subsequent transitional block processes the output of the previous block, with Batch normalization, ReLu, Convolutions of 1x1, dropout and a pool of 2x2 to reduce the input dimensionality and enable the downsampling process. The selected model is pre-trained with Imagenet. 

In order to calculate the model loss we employ the Cross entropy loss between the one-hot encoded vector of true labels $(y)$ of each image and the vector of predicted class probabilities ($\hat{y}$), and $C$ is the number of classes :

\[
{L}(y,\hat{y})= - \sum_{i=1}^{C} y_{i} log(\hat{y_{i}})
\]

While training, this loss will be back-propagated to update the model parameters using Adam Optimizer. The CE loss penalizes incorrect predictions more harshly than correct ones, and encourages the model to assign high probabilities to the correct classes.

This baseline model was compared with EfficientNetV2, which has been shown to achieve state-of-the-art performance on several computer vision benchmarks \cite{tan2021efficientnetv2}. EfficientNetV2 builds upon the previous EfficientNet models by employing training-aware neural architecture search to optimize the network topology, and compressed global memory. The main goal of this model is to optimize training speed and parameter efficiency. Also, this time the search space also included new convolutional blocks such as Fused-MBConv. 

On the other hand, ViT (Vision Transformer) was first introduced as a novel architecture that applied the Transformer architecture, originally designed for natural language processing, to computer vision tasks \cite{dosovitskiy2021image}. It treats images as patches, and use self-attention mechanisms to capture relationships between these patches. 

EfficientNetV2 and ViT are both state-of-the-art models that have been successfully applied to multi-classification tasks in the computer vision literature which is one of the main reasons to compare it with our baseline model. 

\subsection{Evaluation Metrics}

We use the following metrics for evaluating the results of our model as this have been used under the same modality for image classification: 1) accuracy, 2) precision, 3) recall and 4) f-1 score. These metrics let us compare our results with state of the art on the same CheXPert database \cite{DBLP}. 

\begin{enumerate}
    \item Accuracy: Ratio of number of correct predictions to the total number of input samples. It is calculated as: 
        $\text{Accuracy} = \frac{\text{TP} + \text{TN}}{\text{TP} + \text{FP} + \text{TN} + \text{FN}}$

    \item Precision: Provides the proportion of texts classified correctly, so it allows us to know of the number of well-predicted images, what percentage is truly positive according to the labels. High precision minimizes false positives. For each class $C_i$ this will be calculated in the form:$\text{PPV}(C_i) = \frac{\text{TP}(C_i)}{\text{TP}(C_i) + \text{FP}(C_i)}$

    \item Recall: Provides the proportion of well-identified annotations, which will allow us to know about the total number of positives what percentage is correctly classified for each class. A high recall minimizes false negatives. For each class this will be calculated in the form: $\text{TPR}(C_i) = \frac{\text{TP}(C_i)}{\text{TP}(C_i) + \text{FN}(C_i)}$

    \item F-1: The F1 measure is a harmonic average of precision and recall that favors a good balance between both measures. For each class $C_i$ the following is calculated:
        $F_1(C_i) = \frac{2 \cdot (\text{TPR}(C_i) \cdot \text{PPV}(C_i))}{\text{TPR}(C_i) + \text{PPV}(C_i)}$

Where TP is true positive, FP is false positive, FN is false negative and TN is true negative. Higher values of the metrics indicate better results.
    
\end{enumerate}

\section{Dataset}

CheXpert is a large and public dataset for chest radiograph interpretation that consists of 224,316 chest radiographs of 65,240 patients. The radiographic examinations were collected from Stanford Hospital and they were performed between October 2002 and July 2017 in both inpatient and outpatient centers, along with their associated radiology reports.

Each report has a label indicating the presence of 14 observations as positive, negative or uncertain. In the dataset a \textbf{blank} label stands for unmentioned, \textbf{0} for negative, \textbf{-1} for uncertain and \textbf{1} for positive. The 14 observations are: No Finding which means patient is healthy, Support Devices which means patient has a device such as a pacemaker, and 12 diseases or conditions. These thorax diseases are: Enlarged Cardiomediastinum, Cardiomegaly, Lung Lesion, Lung Opacity, Edema, Consolidation, Pneumonia, Atelectasis, Pneumothorax, Pleural Effusion, Pleural Other and Fracture. 

\begin{table}[H]
\centering
\caption{The CheXpert dataset consists of 14 labeled observations. This table reports the number of studies which contain these
observations in the training set.}
\label{Dataset specifications}
\resizebox{0.45\textwidth}{!}{%
\begin{tabular}{|l|c|c|c|}
\hline
\multicolumn{1}{|c|}{\textbf{Pathology}} & \textbf{Positive (\%)} & \textbf{Uncertain (\%)} & \textbf{Negative (\%)} \\ \hline
\textbf{No Finding}                      & 16627 (8.86)           & 0 (0.0)                 & 171014 (91.14)         \\ \hline
\textbf{Enlarged Cardiom.}               & 9020 (4.81)            & 10148 (5.41)            & 168473 (89.78)         \\ \hline
\textbf{Cardiomegaly}                    & 23002 (12.26)          & 6597 (3.52)             & 158042 (84.23)         \\ \hline
\textbf{Lung Lesion}                     & 6856 (3.65)            & 1071 (0.57)             & 179714 (95.78)         \\ \hline
\textbf{Lung Opacity}                    & 92669 (49.39)          & 4341 (2.31)             & 90631 (48.3)           \\ \hline
\textbf{Edema}                           & 48905 (26.06)          & 11571 (6.17)            & 127165 (67.77)         \\ \hline
\textbf{Consolidation}                   & 12730 (6.78)           & 23976 (12.78)           & 150935 (80.44)         \\ \hline
\textbf{Pneumonia}                       & 4576 (2.44)            & 15658 (8.34)            & 167407 (89.22)         \\ \hline
\textbf{Atelectasis}                     & 29333 (15.63)          & 29377 (15.66)           & 128931 (68.71)         \\ \hline
\textbf{Pneumothorax}                    & 17313 (9.23)           & 2663 (1.42)             & 167665 (89.35)         \\ \hline
\textbf{Pleural Effusion}                & 75696 (40.34)          & 9419 (5.02)             & 102526 (54.64)         \\ \hline
\textbf{Pleural Other}                   & 2441 (1.3)             & 1771 (0.94)             & 183429 (97.76)         \\ \hline
\textbf{Fracture}                        & 7270 (3.87)            & 484 (0.26)              & 179887 (95.87)         \\ \hline
\textbf{Support Devices}                 & 105831 (56.4)          & 898 (0.48)              & 80912 (43.12)          \\ \hline
\end{tabular}%
}
\end{table}

For this project, we reduced the dataset. Since our main interest is to classify a chest radiograph among the existing conditions, disregarding uncertainty or negativity and solely identifying positive cases. On the other hand, since the dataset is unbalanced, we aim to balance it and for this we will only select 5 classes. The selected classes are: No Finding, Cardiomegaly, Edema, Pneumothorax and Pleural Effusion. From each pathology/condition we are going to take 8,193 radiographs for the training set. Finally, we have 5,595 chest radiographs for the validation set and 5,595 chest radiographs for the test set.

The selection of the five classes—No Finding, Cardiomegaly, Edema, Pneumothorax, and Pleural Effusion—was based on their prevalence in the dataset, clinical relevance, and the diversity of conditions they represent. These classes have a significant number of positive cases, providing a robust dataset for training the model. No Finding is crucial for establishing a baseline, while Cardiomegaly, Edema, Pneumothorax, and Pleural Effusion represent critical conditions requiring prompt diagnosis and management. The higher number of samples in these categories is not only due to their frequency in the dataset but also reflects their global prevalence \cite{who2019global}. Cardiovascular diseases, pulmonary conditions, and pleural abnormalities are among the most common health issues worldwide, necessitating their prioritization in medical imaging research. For instance, lower respiratory infections and chronic obstructive pulmonary disease (COPD) are among the top global causes of both death and disability-adjusted life years (DALYs) according to the World Health Organization \cite{who2021global}, highlighting the importance of accurately diagnosing these conditions. 

This approach ensures the development of a comprehensive classifier capable of aiding in the accurate diagnosis of these prevalent and significant chest pathologies, ultimately improving patient care.\cite{WangPeng} 

\section{Analysis and Results}

The following results have these hyperparameters: batch\_size of 18, com\_round of 3, 10 epochs, eps of 1e-08, gamma of 0.5, image size of 384x384, log\_interval of 10, loss is CrossEntropyLoss, learning rate of 0.01, momentum of 0.5, no-cuda is False, patience is 5, seed is 22, step\_size is 2, val\_inter of 1, and weight\_decay of 0.0001. On the other hand, the hyperparameter we varied was the learning rate with values of 0.01 and 0.0001. Using this two learning rates we varied the models using DenseNet121, EfficientNet and ViT. 

Tables \ref{EN 0.01} and \ref{EN 0.0001} show the results using the model EfficientNet. As can be seen according to the f1-score, in this case, a lower learning rate has best performance. In detail, the class with best performance metrics is Pleural Effusion with an f1-score of 0.76, followed by No Finding with 0.73.

\begin{table}[H]
\centering
\caption{Classification Report for EfficientNet with learning rate of 0.01}
\label{EN 0.01}
\begin{tabular}{|l|c|c|c|}
\hline
\multicolumn{1}{|c|}{\textbf{Pathology}} & \textbf{Precision} & \textbf{Recall} & \textbf{f-1} \\ \hline
\textbf{No Finding} & 0.47 & 0.66 & 0.55 \\ \hline
\textbf{Cardiomegaly} & 0.14 & 0.22 & 0.17 \\ \hline
\textbf{Edema} & 0.25 & 0.53 & 0.34 \\ \hline
\textbf{Pneumothorax} & 0.18 & 0.33 & 0.23 \\ \hline
\textbf{Pleural Effusion} & 0.58 & 0.03 & 0.07 \\ \hline
\end{tabular}
\end{table}

\begin{table}[H]
\centering
\caption{Classification Report for EfficientNet with learning rate of 0.0001}
\label{EN 0.0001}
\begin{tabular}{|l|c|c|c|}
\hline
\multicolumn{1}{|c|}{\textbf{Pathology}} & \textbf{Precision} & \textbf{Recall} & \textbf{f-1} \\ \hline
\textbf{No Finding} & 0.73 & 0.74 & 0.73 \\ \hline
\textbf{Cardiomegaly} & 0.54 & 0.45 & 0.49 \\ \hline
\textbf{Edema} & 0.50 & 0.63 & 0.56 \\ \hline
\textbf{Pneumothorax} & 0.52 & 0.60 & 0.56 \\ \hline
\textbf{Pleural Effusion} & 0.81 & 0.72 & 0.76 \\ \hline
\end{tabular}
\end{table}

Tables \ref{DN 0.01} and \ref{DN 0.0001} show the results using the model DenseNet121. As it can be seen, a lower learning rate improves the model's performance. Here the classes with better performance metrics are also Pleural Effusion and No Finding with an f1-score of 0.72 each.

\begin{table}[H]
\centering
\caption{Classification Report for DenseNet121 with learning rate of 0.01}
\label{DN 0.01}
\begin{tabular}{|l|c|c|c|}
\hline
\multicolumn{1}{|c|}{\textbf{Pathology}} & \textbf{Precision} & \textbf{Recall} & \textbf{f-1} \\ \hline
\textbf{No Finding} & 0.37 & 0.83 & 0.51 \\ \hline
\textbf{Cardiomegaly} & 0.14 & 0.12 & 0.13 \\ \hline
\textbf{Edema} & 0.29 & 0.04 & 0.07 \\ \hline
\textbf{Pneumothorax} & 0.18 & 0.10 & 0.13 \\ \hline
\textbf{Pleural Effusion} & 0.58 & 0.52 & 0.55 \\ \hline
\end{tabular}
\end{table}

\begin{table}[H]
\centering
\caption{Classification Report for DenseNet121 with learning rate of 0.0001}
\label{DN 0.0001}
\begin{tabular}{|l|c|c|c|}
\hline
\multicolumn{1}{|c|}{\textbf{Pathology}} & \textbf{Precision} & \textbf{Recall} & \textbf{f-1} \\ \hline
\textbf{No Finding} & 0.70 & 0.74 & 0.72 \\ \hline
\textbf{Cardiomegaly} & 0.41 & 0.65 & 0.50 \\ \hline
\textbf{Edema} & 0.53 & 0.60 & 0.56 \\ \hline
\textbf{Pneumothorax} & 0.51 & 0.64 & 0.57 \\ \hline
\textbf{Pleural Effusion} & 0.84 & 0.63 & 0.72 \\ \hline
\end{tabular}
\end{table}

Finally, tables \ref{ViT 0.01} and \ref{ViT 0.0001} show the results using the model ViT. As can be seen, the model with learning rate of 0.01 had a bad outcome, all the metrics had a value of 0. On the other hand, a learning rate of 0.0001 did show a significant improvement. The classes with better performance metrics were No Finding and Pleural Effusion with a f1-score of 0.65 and 0.62, respectively. 

\begin{table}[H]
\centering
\caption{Classification Report for ViT with learning rate of 0.01}
\label{ViT 0.01}
\begin{tabular}{|l|c|c|c|}
\hline
\multicolumn{1}{|c|}{\textbf{Pathology}} & \textbf{Precision} & \textbf{Recall} & \textbf{f-1} \\ \hline
\textbf{No Finding} & 0.00 & 0.00 & 0.00 \\ \hline
\textbf{Cardiomegaly} & 0.00 & 0.00 & 0.00 \\ \hline
\textbf{Edema} & 0.00 & 0.00 & 0.00 \\ \hline
\textbf{Pneumothorax} & 0.00 & 0.00 & 0.00 \\ \hline
\textbf{Pleural Effusion} & 0.00 & 0.00 & 0.00 \\ \hline
\end{tabular}
\end{table}

\begin{table}[H]
\centering
\caption{Classification Report for ViT with learning rate of 0.0001}
\label{ViT 0.0001}
\begin{tabular}{|l|c|c|c|}
\hline
\multicolumn{1}{|c|}{\textbf{Pathology}} & \textbf{Precision} & \textbf{Recall} & \textbf{f-1} \\ \hline
\textbf{No Finding} & 0.67 & 0.64 & 0.65 \\ \hline
\textbf{Cardiomegaly} & 0.34 & 0.50 & 0.41 \\ \hline
\textbf{Edema} & 0.40 & 0.57 & 0.47 \\ \hline
\textbf{Pneumothorax} & 0.35 & 0.45 & 0.40 \\ \hline
\textbf{Pleural Effusion} & 0.76 & 0.52 & 0.62 \\ \hline
\end{tabular}
\end{table}

Overall, a lower learning rate (0.0001) consistently led to better performance across all models; meanwhile, a higher learning rate (0.01) caused the models, especially ViT, to fail in learning meaningful patterns. This could have been because a high learning rate can cause the model to make large updates to the weights, leading to instability in the training process. This results in the model failing to converge properly or overfitting the training data without generalizing well to the validation data. This might lead to the loss function oscillating or diverging rather than converging to a minimum, preventing the model from learning the underlying patterns in the data. Lower learning rates are particularly useful for fine-tuning pre-trained models, which may already have good initial weights that only need slight adjustments. Also lower learning rates allow for more stable and gradual updates to the weights, leading to a smoother convergence process, this helps to better learn and generalize from the training data.

Additionally, DenseNet121 and EfficientNetV2 had the best performance in comparison to ViT. ViT showed the most drastic improvement when switching to a lower learning rate, but it still lagged behind in its performance, overall. The best model was DenseNet121 and/or EfficientNet with a learning rate of 0.0001. This is because both are well designed architectures that perform well on a variety of image classification tasks. They have mechanisms like dense connections (in DenseNet) and compound scaling (in EfficientNet) that help capturing intricate patterns in the data. On the other hand, ViT models are sensitive to hyperparameter settings, so they may require more careful tuning to perform well and they are known to require larger datasets and longer training times to achieve good performance. They may not perform well with high learning rates or smaller datasets due to their reliance on self-attention mechanisms, which need more data to generalize effectively.

\section{Conclusions}
The observed results are largely driven by the stability and effectiveness of training processes. A lower learning rate of 0.0001 provided more controlled and gradual update mechanisms, allowing the models to converge better and generalize from the training data. EfficientNet and DenseNet121 performed well due to their robust architectures and pre-trained weights, while ViT's performance highlighted the need for careful tuning and potentially more data for optimal results. 

For future work, additional tuning of other hyperparameters, such as batch size, and experimenting with more complex learning rate schedules might further improve model performance. Data augmentation techniques and larger datasets could also help enhance model accuracy using images of different resolutions. On the other hand, in the iteration it would be helpful to validate the proposed solution with experts. 
{\small
\bibliographystyle{ieee_fullname}
\bibliography{egbib}
}

\end{document}